\documentclass[runningheads]{llncs}
\makeindex          
\usepackage{graphicx}
\usepackage{hyperref}
\usepackage{wrapfig}
\usepackage{colortbl}
\usepackage[USenglish,english,american]{babel}
\usepackage[english]{babel}
\usepackage{booktabs}
\usepackage[utf8]{inputenc}
\usepackage{times}
\usepackage{multirow}
\usepackage{algorithm,algpseudocode}
\include{pythonlisting}
\makeatletter
\def\BState{\State\hskip-\ALG@thistlm}
\makeatother

\usepackage{multirow}
\usepackage{wrapfig,lipsum,booktabs}

\usepackage{amsfonts}
\usepackage{amsmath}
\usepackage{amssymb}
\usepackage{booktabs}
\usepackage{caption}
\usepackage{cite}
\usepackage{comment}
\usepackage{float}
\usepackage[utf8]{inputenc}
\usepackage{subfig}
\usepackage{adjustbox}
\usepackage{todonotes}
\usepackage{wrapfig}
\usepackage{rotating}

\def\post#1{\ensuremath{{#1}\kern-.05ex\bullet}\,}
\usepackage{xcolor}

\usepackage{tikz}
\usetikzlibrary{arrows,backgrounds,calc,decorations.pathmorphing,decorations.pathreplacing,fit,matrix,patterns,petri,positioning, shapes}

\renewcommand*\abstractname{abstract}

\usepackage{diagbox}

\begin{document}
	\mainmatter              

	\title{Credit Card Fraud Detection Using Asexual Reproduction Optimization}
	\titlerunning{Credit Card Fraud Detection Using Asexual Reproduction Optimization}  %

\author{Anahita Farhang Ghahfarokhi \inst{1} \and Taha Mansouri\inst{2} \and Mohammad Reza Sadeghi Moghadam\inst{1}* \and    Nila Bahrambeik\inst{1} \and Ramin Yavari\inst{3} \and Mohammadreza Fani Sani\inst{4} }

\authorrunning{Anahita Farhang Ghahfaroki et al.} 
\institute{
	$^1$Department of Production and Operation Management, Faculty of Management, University of Tehran, Tehran, Iran\\
	$^2$School of Science, Engineering and Environment, University of Salford, United Kingdom\\ 
	$^3$Department of Information Technology, Duisburg-Essen University, Duisburg, Germany\\ 
	$^4$Process and Data Science Chair, 
	RWTH Aachen University, Aachen, Germany\\
	\email{anhta.farhang@ut.ac.ir, t.mansouri@salford.ac.uk, 
	rezasadeghi@ut.ac.ir, 
	nilabahrambeik@ut.ac.ir,
	ramin.yavari@stud.uni-due.de,
    fanisani@pads.rwth-aachen.de}}

\maketitle        
   	
\begin{abstract}
As the number of credit card users has increased, detecting fraud in this domain has become a vital issue. Previous literature has applied various supervised and unsupervised machine learning methods to find an effective fraud detection system. However, some of these methods require an enormous amount of time to achieve reasonable accuracy. In this paper, an Asexual Reproduction Optimization (ARO) approach was employed, which is a supervised method to detect credit card frauds. ARO refers to a kind of production in which one parent produces some offspring. By applying this method and sampling just from the majority class, the classification's effectiveness is increased. A comparison to Artificial Immune Systems (AIS), which is one of the best methods implemented on current datasets, has shown that the proposed method is able to remarkably reduce the required training time and at the same time increase the recall that is important in fraud detection problems. The obtained results show that ARO achieves the best cost in a short time, and consequently, it can be considered as a real-time fraud detection system.

\keywords {Machine Learning \(\cdot\) Asexual Reproduction Optimization\(\cdot\) Credit Card Fraud Detection \(\cdot\) Fraud Detection  \(\cdot\) Artificial Immune Systems. }
	\end{abstract}

	\section{Introduction}
	\label{sec:intro}

 Credit card fraud inflicts plenty of costs on banks and card issuers and threatens their reputation \cite{Halvaiee2014novel}. A huge amount of money disappears annually from legitimate accounts by fraudulent transactions \cite{Dal_Pozzolo2015Credit}.
 In fact, E-business has become one of the most important global markets which demands strong fraud detection systems \cite{vsumak2017investigation,Bhattacharyya2011Data}. In 2017, Online Fraud Report of Cyber Source distinguishes average annual fraud loss among different order channels\footnote{http://www.cybersource.com}. 0.9\% of the annual e-commerce revenues is lost due to payment frauds through Web store channel in North America. This value is 0.8\% for Mobile channels and 0.3\% for phone/mail order channel.
 Different definitions of fraud have been presented by different organizations. Based on The World Bank Group’s definition of fraud, the fraudulent practice covers solicitation, offering or taking bribes, or the manipulation of loans in the form of misrepresentation \cite{aguilar2000preventing}. According to the division of the Association of Certified Fraud Examiners, there are two types of fraud, i.e., internal frauds and external frauds. Internal fraud occurs when an employee deliberately misuses an organization’s properties \cite{phua2010comprehensive}. External frauds include a more comprehensive variety in comparison with internal frauds. Dishonest vendors who take bribes are a desirable example to mention. Untruthful customers might alter account information to mislead payments. Besides, third parties may use intellectual properties \cite{chen2013novel}.

The credit card fraud techniques have changed over time, from physically stealing the cards to online frauds \cite{Bhattacharyya2011Data}. Credit card frauds are categorized into two categories, i.e., application frauds and behavioral frauds. An application defrauder is a person who gets a new credit card from issuing companies by utilizing the wrong information. A behavioral defrauder is a person who has attained the information of a legitimate card fraudulently and makes purchases when the cardholder is not present \cite{bolton2001unsupervised}.
As the number of frauds increases, the fighting techniques against fraud become more significant \cite{kou2004survey}. Protection techniques against fraud include prevention and detection systems. The first layer to protect the system against fraud is prevention. Fraud prevention stops the fraud from occurring at the initial level. Fraud detection is the next protection step. It identifies fraudulent activities when they penetrate the system~\cite{behdad2012nature}. 
 People use credit card-based online payments more and more these days, forcing the banks to deploy fraud detection systems \cite{fiore2019using}. Expert-driven, data-driven, and the combination of both are the three kinds of fraud detection systems. Expert-driven systems are based on fraud scenarios. If the data-stream matches the scenario from the FDS viewpoint, the fraud has happened. Data-driven methods learn the fraud patterns and find them in data streams \cite{Jurgovsky2018Sequence}.

Credit card fraud happens when a transaction on someone's credit card is done by another person \cite{tripathi2013hybrid}. If the fraud becomes a prevalent issue in a competitive environment without any preventive systems, it will threaten businesses and organizations seriously \cite{phua2010comprehensive}. On the other hand, the number of credit card transactions is increasing rapidly, which results in the growth of fraudulent activities \cite{zareapoor2012analysis}  It is pretty expensive to analyze the transaction is done by the client or not \cite{Gadi2008Credit}. The fraud detection system is aimed to stop it as soon as possible. Whether the fraud detection system is manual or automatic, it has to be effective. The system should identify a high percentage of fraudulent transactions while keeping the false alarm rate low. Otherwise, the users will become apathetic to alarms \cite{axelsson2000base}. To reduce the cost of detection, many machine learning techniques have been implemented. Supervised methods are more common than unsupervised techniques \cite{bolton2001unsupervised}.

Nowadays, different data mining techniques have been developed \cite{ghahfarokhi2022clustering,ghahfarokhi2021process, berti2022scalable,ghahfarokhi2021python, rohrer2022predictive, grid/FarhangS20} and by acknowledging the development of data mining methods, efficient ways have been found to detect fraud \cite{Carneiro2017data}. However, many of these methods need a time-consuming training phase. This limitation decreases the applicability of these methods. To address this problem, we propose to use Asexual Reproduction Optimization (ARO). In this paper, we implemented and applied this method on a publicly available dataset. The experimental results show that using the proposed method enables us to achieve reasonable accuracy faster, compared to one of the state-of-the-art fraud detection methods, i.e., Artificial Immune Systems (AIS).

 The remaining part of the paper has been organized as follows. \autoref{sec:back} provides a literature review on credit card fraud detection methods. Following, \autoref{sec:mat}  describes the ARO and AIS models. Afterwards, experimental results are presented in \autoref{sec:experiments} and analyzed in \autoref{sec:disc}. Finally, \autoref{sec:conc} concludes the paper and provides some new directions to continue this research.

 	\section{Credit Card Fraud Detection Methods}
	\label{sec:back}

 \begin{table}[tb]
	\caption{Approaches for credit card fraud detection. }
	\resizebox{0.99\textwidth}{!}{
	\begin{tabular}{lll}
		\hline
		Machine   learning algorithm & Method & references \\ \hline
		\multirow{14}{*}{Supervised classifying} & k-Nearest Neighbors (KNN) & \cite{Awoyemi2017Credit}, \cite{Dhankhad2018Supervised},   \cite{Halvaiee2014novel}, \cite{Kumari2019Analysis}, \cite{Yu2009Research},
		\cite{itoo2020comparison},
		\cite{prusti2019credit},
		\cite{bagga2020credit},
		\cite{Zareapoor2015Application} \\ \cline{2-3}
		& Bayesian   Networks (BN)&  \cite{Sa2018customized},    \cite{Filippov2008Credit}, \cite{Gadi2008Comparison},\cite{Gadi2008Credit},\cite{Gadi2008Credit}, \cite{Maes2002Credit}, \cite{Panigrahi2009Credit}, 
		\cite{taha2020intelligent},
		\cite{Yee2018Credit} \\ \cline{2-3}
		& Decision   Trees (DT) & \cite{Devi2017Fraud}, \cite{Dhankhad2018Supervised},  \cite{Gadi2008Comparison}, \cite{Gadi2008Credit}, \cite{Kavitha2017Hybrid}, \cite{Kokkinaki1997On}, \cite{Kumari2019Analysis}, \cite{Minegishi2011Proposal2} \cite{Patil2010Efficient}   \cite{Randhawa2018Credit} \cite{Sahin2011Detecting};  \cite{Sisodia2017Performance}, 
		\cite{husejinovic2020credit},
		\cite{hammed2020implementation},
		\cite{singh2019adaptive},
		\cite{lenka2020investigation},
		\cite{Zareapoor2015Application}\\ \cline{2-3}
		& Artificial Immune Systems (AIS) & \cite{Brabazon2010Identifying}, \cite{Gadi2008Comparison}, \cite{Gadi2008Credit}, \cite{Halvaiee2014novel}, \cite{Tuo2004Artificial}, \cite{Wong2012Artificial}\\ \cline{2-3}
		& Naïve Bayes (NB) & \cite{Akila2017Credit}, \cite{Akila2017Risk}, \cite{Alowais2012Credit}, \cite{Awoyemi2017Credit}, \cite{Dhankhad2018Supervised},  \cite{Filippov2008Credit}, \cite{Gadi2008Credit}, \cite{Gadi2008Comparison}, \cite{Kumari2019Analysis}, \cite{Mohammed2018Scalable}, \cite{Monika2018Fraud}, \cite{Randhawa2018Credit},  \cite{Yee2018Credit},
		\cite{bagga2020credit},
		\cite{bagga2020credit},
		\cite{itoo2020comparison},
		\cite{varmedja2019credit},
		\cite{singh2019adaptive},
		\cite{prusti2019credit},
		\cite{husejinovic2020credit},
		\cite{Zareapoor2015Application}\\ \cline{2-3}
		& Support   Vector Machines (SVM) & \cite{Bhattacharyya2011Data}, \cite{Carneiro2017data}, \cite{Chen2006new}, \cite{Chen2004Detecting}, \cite{Chen2005Personalized}, \cite{Chen2005Novel}, \cite{Devi2017Fraud}, \cite{Dhankhad2018Supervised},  \cite{Kumari2019Analysis}, \cite{Lu2011Research},
		\cite{Randhawa2018Credit},   
		\cite{Sahin2011Detecting}, \cite{Sisodia2017Performance}, \cite{Tran2018Real},
		\cite{prusti2019credit},
		\cite{Whitrow2009Transaction},
		\cite{makki2019experimental},
		\cite{Wiese2009Credit},  \cite{Zareapoor2015Application}\\ \cline{2-3}
		& Logistic   Regression & \cite{Awoyemi2017Credit},   \cite{Bhattacharyya2011Data}, \cite{Carneiro2017data}, \cite{Dhankhad2018Supervised}  \cite{Randhawa2018Credit}
		\cite{bagga2020credit},
		\cite{itoo2020comparison},
		\cite{makki2019experimental},
		\cite{varmedja2019credit},
		\cite{Sahin2011Detecting}; \cite{Vlasselaer2015APATE}\\ \cline{2-3}
		& Random   Forest & \cite{Alowais2012Credit}, \cite{Bhattacharyya2011Data}, \cite{Carneiro2017data}, \cite{Devi2017Fraud}, \cite{Dhankhad2018Supervised},   \cite{Jurgovsky2018Sequence}, \cite{Kumari2019Analysis} \cite{Mohammed2018Scalable}, \cite{Randhawa2018Credit}, \cite{Sohony2018Ensemble}, \cite{Vlasselaer2015APATE},
		\cite{lenka2020investigation},
		\cite{sailusha2020credit},
		\cite{devi2019cost},
		\cite{singh2019adaptive},
		\cite{varmedja2019credit},
		\cite{kumar2019credit},
		\cite{Whitrow2009Transaction};\cite{Xuan2018Random};   
		\cite{lucas2020towards}\\ \cline{2-3}
		& Genetic   Algorithm (GA) & \cite{Duman2011Detecting}, \cite{Ma2009Application}, \cite{Monirzadeh2018Detection}, \cite{Ozcelik2010Improving}, \cite{Patel2013Credit}, \cite{RamaKalyani2012Fraud}, \cite{Wu2007real-valued}\\ \cline{2-3}
		&  Neural   Networks (NN) & \cite{Aleskerov1997Cardwatch}, \cite{Behera2017Credit}, \cite{Brause1999Neural}, \cite{Carneiro2015Cluster}, \cite{Dal_Pozzolo2015Credit}, \cite{Dhankhad2018Supervised},  \cite{Dorronsoro1997Neural}, \cite{fiore2019using}, \cite{Fu2016Credit}, \cite{Gadi2008Comparison}, \cite{Gadi2008Credit}, \cite{Ghosh1994Credit}, \cite{Gomez2018End-to-end},
		\cite{arya2020deal},
		\cite{cheng2020spatio},
		\cite{bagga2020credit},
		\cite{dubey2020credit},
		\cite{carrasco2020evaluation},
		\cite{makki2019experimental},
		\cite{gangwar2019wip},
		\cite{Gulati2017Credit},
		\cite{Guo2008Neural},
		\cite{kolli2020fraud},
		\cite{Jog2018Implementation},\\
		& &  
			\cite{Jurgovsky2018Sequence}, \cite{Khan2014Real-time}, \cite{Kim2002neural}, \cite{Kumari2019Analysis}, \cite{Maes2002Credit}, \cite{Modi2013Fraud},   \cite{Monirzadeh2018Detection}, \cite{Patidar2011Credit},   \cite{Randhawa2018Credit},  \cite{Roy2018Deep}, \cite{Sahin2011Detecting}, \cite{Sohony2018Ensemble}, \cite{Syeda2002Parallel}, \cite{Vlasselaer2015APATE},
			\cite{lebichot2019deep},
			\cite{zhang2019hoba},
			\cite{Wang2018Credit},
			\cite{Wiese2009Credit} \\ \cline{2-3}
		& Scatter   Search & \cite{Duman2011Detecting}\\ \cline{2-3}
		& APATE & \cite{Vlasselaer2015APATE}\\ \cline{2-3}
		& Fisher   Discriminant & \cite{Mahmoudi_2015_Fraud}\\   \hline
		\multirow{5}{*}{Unsupervised clustering} & Self-Organizing   Maps & \cite{Olszewski2014Fraud}, \cite{Quah2008Real-time},  \cite{Zaslavsky2006Credit} \\ \cline{2-3}
		& Fuzzy & \cite{Behera2017Credit}, \cite{Bentley2000Fuzzy}, \cite{Sanchez2009Association}, \cite{Sarno2015Hybrid}\\ \cline{2-3}
		& Principal   Component Analysis & \cite{lepoivre2016credit}\\ \cline{2-3}
		& Hidden   Markov Model (HMM) & \cite{Bhusari2013Detailed}, \cite{Falaki2012Probabilistic}, \cite{Khan2013Credit}, \cite{Kumari2014Credit}; \cite{Rani2011Credit}, \cite{Srivastava2008Credit}, \cite{lucas2020towards}\\ \cline{2-3}
		& Simple   K-Means & \cite{lepoivre2016credit}\\ \hline
	\end{tabular} }
	\label{tab:FD}
\end{table}

Fraud detection merges anomaly-based detection and misuse-based detection by applying data mining techniques. Anomaly-based detection consists of supervised, unsupervised, and semi-supervised algorithms \cite{abdallah2016fraud}. Supervised algorithms require all existing transactions, which are labeled as fraudulent and non-fraudulent transactions. These algorithms assign a score to a new transaction, which determines the transaction's label \cite{bolton2001unsupervised,phua2010comprehensive}. Unsupervised methods work with unlabeled test dataset and try to find the unusual transactions. These algorithms represent a baseline distribution for the normal behavior. Transactions with a great distance from it are considered unusual ones \cite{bolton2001unsupervised,zhu2009introduction}. Semi-supervised methods contain both labeled and unlabeled instances. Semi-supervised learning aims to design the algorithms, which can use these combined instances \cite{zhu2009introduction}. In general, the concept of anomaly/outlier is problem-dependent and it is challenging to capture all aspects of behavior in one single metric \cite{campos2019outlier}. In \autoref{tab:FD}, we presented some of the data mining based approaches which are used for credit card fraud detection, carried out in the literature \cite{abdallah2016fraud,kultur2017novel,west2016intelligent,singh2020empirical}.

In \cite{Aleskerov1997Cardwatch}, the authors presented a neural network-based system with a user-friendly interface for fraud detection, implemented on synthetic datasets \cite{Aleskerov1997Cardwatch}. In credit card fraud detection, datasets have skewed distributions. Chen et al. employed Binary Support Vector System (BSVS), which could handle this problem better than oversampling techniques. For support vector selection, the genetic algorithm is used. Based on these vectors, they proposed BSVS \cite{Chen2006new}. Gadi et al. employed BN, NB, AIS, and DT techniques on the Brazilian bank dataset that we used in this paper. They showed that generally applying cost-sensitive and robust optimization leads to better results \cite{Gadi2008Comparison}. Because of optimizing the parameters, AIS is the best technique \cite{Gadi2008Credit}. Sánchez et al. applied the association rules in the credit card fraud detection system. The system determined patterns for legitimate transactions. The transactions that do not match with the patterns are recognized as fraudulent \cite{Sanchez2009Association}. Instead of looking individually at data, authors in \cite{Wiese2009Credit} consider them sequentially. They applied SVM and Long Short-Term Memory Recurrent Neural Network (LSTM) for modeling time series in fraud detection records. LSTM was a more suitable classifier \cite{Wiese2009Credit}. Rani et al. suggested a method using HMM which could conserve user’s data effectively and bring back the information with ease \cite{Rani2011Credit}. Modi et al. examined a single-layer feed-forward neural network for fraud detection. The fraud categorization was divided into four groups of low to high risk. If a transaction is recognized as a fraudulent one, it will belong to one of these groups \cite{Modi2013Fraud}.

Using negative selection in addition to clonal selection, Halvaiee and Akbari improved AIS. They suggested a new method AIS-based Fraud Detection Model (AFDM) for calculating the samples' fitness. Furthermore, in their proposed model, they used cloud computing for training, which reduced the processing time \cite{Halvaiee2014novel}. Zareapoor and Shamsolmoali examined bagged ensemble decision tree on a real dataset and compared it with SVM, KNN, and NB. It achieved the highest detection rate. The time was reduced significantly, and the ensemble technique could solve the imbalanced dataset problem \cite{Zareapoor2015Application}. Carneiro et al. aim the development and implementation of a fraud detection system at an e-tail merchant. They showed that choosing the right variables in the dataset is a key factor. Random forests, logistic regression, and support vector machines were tested. A random forest can be an appropriate practical model \cite{Carneiro2017data}. Fiore et al. used Generative Adversarial Networks (GAN) to detect credit card fraud. GAN is a multiple-layer neural network consisted of a generator and a discriminator. They employed GAN for solving imbalanced dataset problem. GAN generates an augmented dataset that has more fraudulent transactions than the initial dataset \cite{fiore2019using}. Behera and Panigrahi proposed a two-stage system. The first stage tries to match the patterns. 
 It consists of a fuzzy module which computes a score. Given this score, one can envisage three categories: legitimate, fraudulent, and suspicious. The next stage concludes a neural network, which determines whether the suspicious one belongs to a fraudulent or legitimate group \cite{Behera2017Credit}.
 
 De Sá et al. implemented a customized Bayesian Network Classifier (BNC) on the dataset of a Brazilian payment service. They used a Hyper-Heuristic Evolutionary Algorithm for generating BNC. The proposed method increased economic efficiency remarkably \cite{Sa2018customized}. Gómez et al. used an end-to-end neural network for credit card fraud detection. They focused on solving imbalanced dataset and cost evaluation problems and obtained valuable results\cite{Gomez2018End-to-end}. Lucas et al. modelled a sequence of credit card transactions from three different perspectives. Each of these sequences with HMM and the likelihood associated with HMM is used as additional features in the Random Forest classifier for fraud detection \cite{lucas2020towards}. Gianini et al. used a game theory-based approach for detecting credit card fraud by managing a pool of rules \cite{gianini2020managing}.
 
 Monirzadeh et al. increased the efficiency of the neural network by using the genetic algorithm. Their research showed that the most effective criterion is the information related to the transaction. Age, gender, and such factors do not affect the detection \cite{Monirzadeh2018Detection}.
 In any fraud detection system, the chief problem is always to increase the accuracy of approving a legal transaction, whether in the shortest possible time or at the lowest cost for financial institutions \cite{Gadi2008Credit}. Therefore, the principal purpose of all the models presented for this issue is to reduce the detection time, increase the accuracy, reduce the costs, and present a model that can improve these factors with better performance.
 According to the description, the algorithms' performance has been compared through the three aspects of fraud detection speed, accuracy, and cost presented in \autoref{tab:comp} \cite{zareapoor2012analysis}.
 
 As shown in \autoref{tab:comp}, most of the algorithms have some disadvantages in the mentioned indicators, and among them, AIS performs the best. This confirms the results presented in \cite{Gadi2008Comparison}, where different techniques are compared with each other and AIS is the best technique based on their costs \cite{Gadi2008Credit}. For this reason, it is chosen for comparison with the ARO algorithm. We employed ARO, which is a supervised method for credit card fraud detection. ARO is an asexual reproduction optimization algorithm. Like Particle Swarm Optimization (PSO), Genetic Algorithms (GA), and Ant Colony Optimization (ACO), ARO is also an Evolutionary Single-Objective Optimization technique \cite{mansouri2011aro}.
 ARO has some advantages that make it completely different from other algorithms. First, it is an individually based technique which reaches the global optimal point, astonishingly faster than other algorithms. Thus, unlike population-based algorithms that require a large number of computational resources to convert, ARO consumes much fewer resources and converges faster. The second case is about mathematical convergence. It has good exploration and exploitation rates. Third, ARO does not require parameter settings, so it is unlikely to have trouble in setting parameters, which is a common meta-cognitive problem of Genetic Algorithms (GA), Annealing Simulation (SA), Taboo Search (TS), and Particle Swarm Optimization (PSO). Besides, ARO does not use any selective mechanism such as a roulette wheel. Inappropriate adoption of selection mechanisms may lead to problems such as premature convergence due to excessive selection pressure. Fourth, ARO is a free model algorithm that can be applied to various types of optimization \cite{farasat2010aro,salmeron2019learning}. For all of the above reasons, ARO can be selected for comparison with the AIS in fraud detection problems. ARO has not been used in fraud detection up to this point. In this paper, a comparison is made between ARO and AIS. We run ARO on the same dataset on which the AIS has been implemented \cite{Gadi2008Comparison,Halvaiee2014novel}.

  \begin{table}[tb]
 	\caption{Comparison of algorithms.}\scriptsize
 	\label{tab:comp}
 	\resizebox{\textwidth}{!}{
 	\begin{tabular}{|l|l|l|l|l|l|l|l|l|l|}
 		\hline
 		Algorithm & NN & BN & SVM & KNN & DT & fuzzy & AIS & GA & HMM \\ \hline
 		Fraud   detection speed & Fast & Very fast & Low & Good & Fast & Very low & Very fast & Good & Fast \\ \hline
 		Accuracy & Medium & High & Medium & Medium & Medium & Very high & Good & Medium & Low \\ \hline
 		Cost & Expensive & Expensive & Expensive & Expensive & Expensive & High expensive & Inexpensive & Inexpensive & High expensive \\ \hline
 	\end{tabular}
 	}
 \end{table}
 
	\section{Using ARO for Fraud Detection}
	\label{sec:mat}
In this section, we explain ARO in more details and how we implement that to detect fraud. 
Moreover, the AIS algorithm that has the highest performance is briefly explained~\cite{Halvaiee2014novel}. 
As the ARO method is a supervised method, we need to separate the data into train and validation parts. 
Therefore, like any other supervised method, we use the train part of data for the training and the validation part for the testing phase. 

\subsubsection{ARO algorithm}
ARO is taken from asexual reproduction. Asexual reproduction refers to a kind of production in which one parent produces offspring identical to herself \cite{faust2002identifying}. In populations like bacteria, asexual reproduction is prevalent \cite{holmstrom2004runs}. There are several kinds of asexual reproduction like budding \cite{prall2007high}, asexual spore production \cite{lee2006light}, and binary fission \cite{song2004morphogenesis}. ARO is inspired by the budding method. In the budding method, the parent produces a smaller copy of itself called a bud. The bud separates itself from the parent to become an independent one \cite{mansouri2011aro}.

Here, we explain how we use ARO for detecting frauds.  
According to the label of transactions in the training data, two separate matrices were created for fraud and legal transactions. For each feature in the legal matrix, the maximum and the minimum values are determined and placed in the maximum and the minimum legal matrices, and a parent is created randomly between the values of the maximum and the minimum matrices.
Note that the value of each bit in this parent is a random value between its corresponding bit in the maximum and the minimum legal matrices.
The value of the parent fitting was calculated using the fitting function given in \autoref{equ:e1} and named “parent-fitting”.

\begin{equation}
\label{equ:e1}
distance_{record-normal-transtions}= \frac{\sum_{i=1}^{M}\sum_{j=1}^{N}(\frac{|r_i-nt_{ji}|}{max_i-min_i})}{kN}
\end{equation}
Where $M$ is the number of features in our dataset.

The cut point in a dataset is the best fitness achieved in that dataset.

\begin{equation}
\label{equ:e1}
distance_{record-normal-transtions}= \frac{\sum_{i=1}^{M}\sum_{j=1}^{N}(\frac{|r_i-nt_{ji}|}{max_i-min_i})}{kN}
\end{equation}

Afterwards, we repeat the following process until the parent fit is smaller than the cut point of the data set.

\begin{itemize}
	\item Select the starting bit (S) as a random number within the range of the number of features. Select the end (E) between the starting bit and the last number of features. Calculate the probability of mutation through \autoref{equ:e2}:
	
	\begin{equation}
	\label{equ:e2}
	P=\frac{1}{1+Ln(E-S+1)}
	\end{equation}
	
	\item	Put the bud equal to the parent.
	\item	For the bits between the starting and ending bit selected randomly as above, if the probability P calculated in \autoref{equ:e2} is greater than or equal to an arbitrary random number between zero and one in MATLAB, the value of the bit will be mutated. In this way, the bud will be mutated.
	\item The value of the mutated bud in \autoref{equ:e3} is calculated (using the fitting function) and named as "bud fit".
	\begin{equation}
	\label{equ:e3}
	\resizebox{.94\hsize}{!}{
	$distance_{record{-}final{-}transtions}{=} distance_{record{-}fraud{-}transtions} {-} distance_{record{-}normal{-}transtions}$
	}
	\end{equation}
	\item	If the bud is fitted, it is more than the parent, and the bud replaces the parent. Fitting the bud replaces the parent fitting. The bud is added to the identifier matrix, and one unit is added to the count of the matrix rows of the identifier.
	
	\item In each fitting calculation, separate bud fits are calculated for the fraudulent matrix and the legal matrix. Fit the bud for the fraudulent matrix added to fraudulent matrix \autoref{equ:e4}. One is added to the counter of the fraudulent fitting matrix. The fitting of the bud for the legal matrix is added to the legal matrix \autoref{equ:e1}, and one is added to the count of the legal matrix rows. The loop termination condition reaches to a value more than or equal to the parent fit compared to the cut point.
	\begin{equation}
	\label{equ:e4}
	distance_{record-fraud-transtions}= \frac{\sum_{i=1}^{M}\sum_{j=1}^{F}(\frac{|r_i-ft_{ji}|}{max_i-min_i})}{kF}
	\end{equation}
\end{itemize}
\begin{figure}[htb!]
	\centering
	\includegraphics[width=\textwidth]{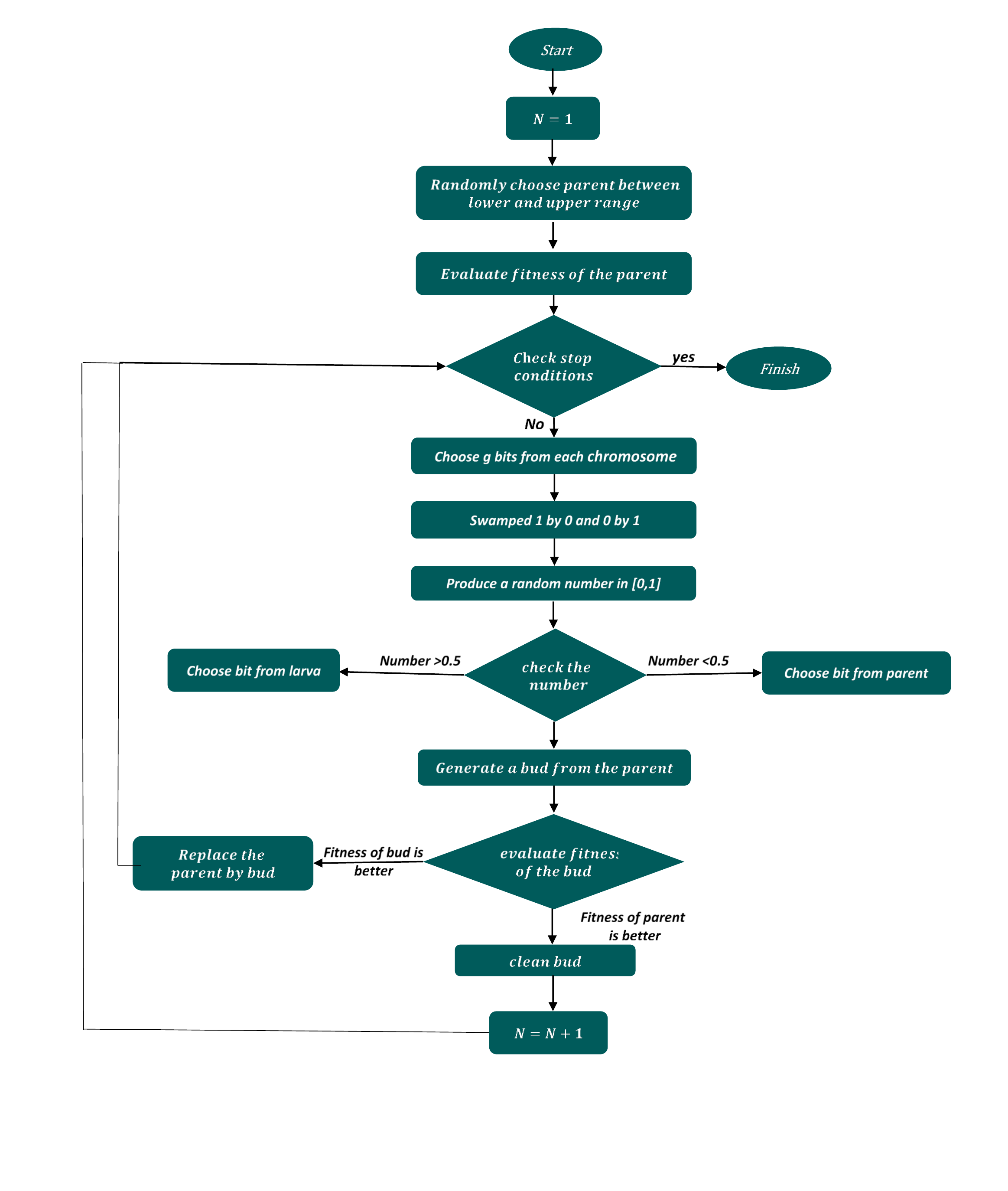}
	\vspace{-1.7cm}
	\caption{Flowchart of ARO.}
	\label{fig:chart}
\end{figure}
The schematic view of ARO algorithm is presented in \autoref{fig:chart}. In ARO algorithm, an individual is shown by a vector of variables $ X {=} (x_1, x_2, … , x_n)$, $X {\in} \mathcal{R}_n$. Each variable is considered as a chromosome. A binary string represents a chromosome consisted of genes. The length of the string is $L {=} l_1 + l_2 + 1$. It is supposed that every generated answer exists in the environment, and because of limited resources, only the best solution can remain alive. The algorithm starts with a random individual in the answer scope. This parent reproduces the offspring named bud. Just the parent or the offspring can survive. In this competition, the one which outperforms in fitness function remains alive. If the offspring has suitable performance, it will be the next parent, and the current parent becomes obsolete. Otherwise, the offspring perishes, and the present parent survives. The algorithm recurs until the stop condition occurs.

\begin{figure}[htb!]
	\centering	
	\includegraphics[width=\textwidth]{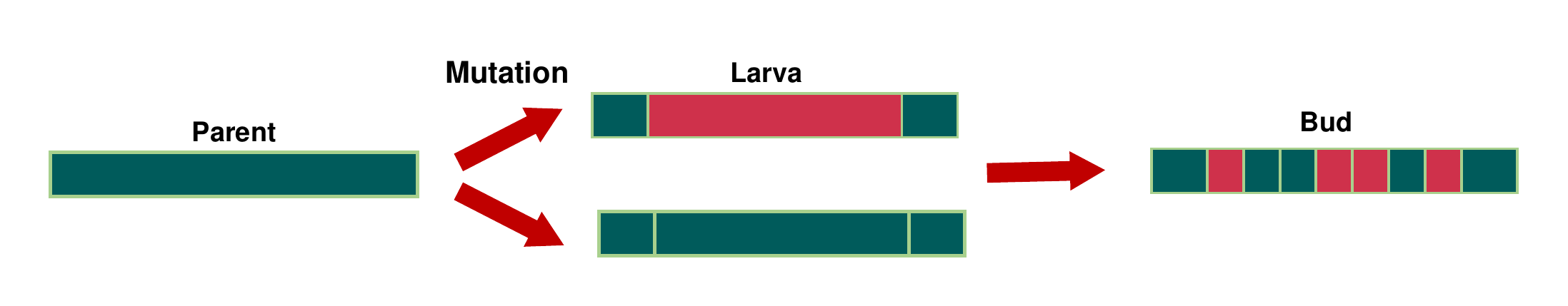}     \caption{Process of bud reproduction.}
	\label{fig:bud}
	
\end{figure}

In the reproduction stage, a substring with $\lambda$bits is picked out in all chromosomes, which is named larva. $\lambda$ is a random number between 1 and L. In the exploration phase, the substring is mutated, in each gene in the substring, 1 is swamped by 0 and 0 by 1. In the exploitation phase, the parent and larva merge as shown in \autoref{fig:bud}. Process of bud reproduction. If $P$ which is calculated from $P{=}\frac{1}{1+{L_n(\lambda)}}$ is higher than $0.5$, the bud gene is picked out from the larva, otherwise the bud gene will be picked out from the parent chromosome.
\begin{figure}[htb!]
	\centering
	\includegraphics[scale=0.5]{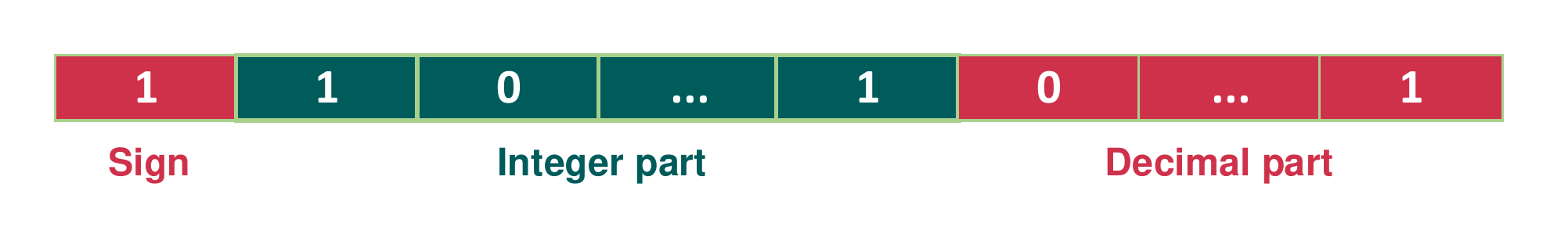}     
	\caption{A model for a chromosome in ARO. }
	\label{fig:AROModel}
\end{figure}	

\autoref{equ:e5} relates the exploitation and exploration. If $\lambda$ is a big number, less exploitation is needed and vice versa. In fact, exploration and exploitation are inversely related.

\begin{equation}
    \label{equ:e5}
    P{=}\frac{1}{1+L_n(\lambda)}
\end{equation}

The fitness of both bud and parent is calculated to choose the best one for the algorithm's next run after reproduction \cite{farasat2010aro, mansouri2011aro}.
Note that we do this procedure for all records and all features. 
Each record has a fraud or normal label. There are the following hints to mention:
\begin{enumerate}
    \item According to \autoref{fig:AROModel}, a chromosome has three parts. Here, just the integer part is considered because we do not have the sign or decimal part.
    \item Genes are not binary and they contain integer numbers.
    \item Only the normal (or legal) records are sampled because the dataset is skewed toward normal transactions. The number of normal transactions is significantly more than the fraudulent transactions. Thus, normal records society is suitable for sampling versus the fraudulent society.
    \item For generating the first parent, one should determine the range for each bit (gene). Then, for each gene in the first parent chromosome, a random number between the maximum and the minimum of that gene is chosen.
    \item The fitness function will be used, which is described in the next section.
    \item First, a larva should be generated when reproducing a bud. For generating a larva, a random length should be created. Each gene in this length assumes a random number between the maximum and the minimum of that gene. This length would be a larva. The next step for reproducing a bud is choosing the gene between larva and parent, like in \autoref{fig:bud} (the process of bud reproduction). In this step, for choosing each gene between larva and parent, a random number is generated. If the random number is less than $p$, which is obtained from (1), the gene is selected from the larva, otherwise the gene is selected from a parent.
\end{enumerate}

Parameter setting causes plenty of problems in methods such as PSO and GA. ARO does not need parameter setting. ARO is an individual-based technique which saves time, unlike population-based techniques that waste time. ARO can be used in different kinds of optimization issues despite many algorithms, which can only be used in one sort of optimization problems.
Adjustment with a diverse genetic environment is one of the problems faced in ARO. However, it can be solved by special reproduction operators \cite{mansouri2011aro}.

\textbf{Fitness function}
In case one decides to evaluate fitness for a specific record, at first, the distance between the record and all fraud transactions is calculated by Equation 4, and then the distance between the record and all normal transactions is calculated by Equation 1. The difference between these two numbers, as shown in Equation 3, would be the fitness. Due to only sampling the normal records, the higher is the fitness number, the better it is because it shows that the record is closer to the normal transactions than the fraud ones. Thus, it can be a suitable normal sample.
In Equations 4 and 1, each record is considered to have $k$ fields. Here, $k$ is 17. The value of the $i$'th field of the record is $r$, the value of the $i$'th field of the \textit{j}'th normal transaction is \textit{$nt_{ji}$}, and the value of the \textit{i}'th field of the \textit{j}'th fraud transaction is \textit{$ft_{ji}$}. The maximum and the minimum of \textit{i}'th field in all records of dataset are represented by \textit{$max_{i}$} and \textit{$min_{i}$}. The number of all normal transactions in the considered dataset is \textit{$N$}, and the number of all fraudulent transactions is \textit{$F$}.

\subsubsection{AIS algorithm}
AIS is inspired by the immune system of the human body. It creates the detectors called lymphocytes for identifying non-self-cells like viruses.
Negative selection and clonal selection are two stages of the AIS. Through negative selection, lymphocytes are created by a random combination of protein patterns. Lymphocytes should not detect self-cells. Thus, the immune system eliminates the lymphocytes that react to self-cells. In fact, all of the lymphocytes generated randomly that react to self-cells are eliminated immediately after creation, and other lymphocytes survive. This procedure is named negative selection.
After negative selection, a short life starts for the remaining lymphocytes. They meet any non-self-cells. If any lymphocyte reacts to a non-self-cell, it can survive to protect the body against those non-self-cells.  This procedure is named clonal selection. The lymphocyte which detects a non-self-cell is cloned by mutation. The colony cells, which are closer to the non-self-cell, are chosen to survive. These colony cells are considered as memory cells and will react to non-self-cells like viruses \cite{Halvaiee2014novel}.

Both non-self-/self-cells are considered as vectors. At first, the training-set should be normalized. Initializing the parameters is the next step. Then, $N_{pop}$ of normal records is selected randomly as primary detectors (Just the normal records like ARO were sampled). The affinity of these records is calculated using the distance function. $N_c$ of the records with higher affinity is selected. A colony is expanded from them. It means the records with more affinity will be replicated more. The colony is mutated. $N_{m}$ of the best-mutated population is chosen to replicate $N_{m}$ of the worst memory cells. This algorithm continues until the stop condition occurs \cite{Halvaiee2014novel}. Here, the loop repeats are considered 150 times. $N_{pop}$, $N_c$, and $N_m$ are 25, 7, and 5, which have been driven from Gadi et al. and Halvaiee \cite{Gadi2008Credit,Halvaiee2014novel}.

\begin{center}
     \colorbox[RGB]{232,242,225}{
         \begin{minipage}{1\linewidth}

              \begin{algorithm}[H]
              	\caption{AIS}
	\label{pseudoPSO}
	\begin{algorithmic}[1]
		\State  Determine  $N_{pop}$  $\%$ the number of all detectors
		\State Determine  $N_{c}$  $\%$ the number of detectors best match with non - self cell
		\State Determine  $N_{m}$ $\%$ the number of best mutated detectors
		\While {stop conditions do not occur} 
		\State Choose $N_{pop}$ of population randomly, call it $first{-}pop$
		\State Choose $N_c$ of best $first{-}pop$ based on their fitness, call it $best{-}first{-}pop$
		\State Expand a colony from $best{-}first{-}pop$, call it $colony{-}pop$
		\State Mutate $colony{-}pop$, call it $mutated{-}pop$
		\State Choose $N_m$ of best $mutates{-}pop$ based on their fitness, call it $best{-}mutated{-}pop$;
		\State   Replace $N_m$ of worst detectors in memory cell by $best-mutated-pop$
		\EndWhile
			\end{algorithmic}
               \end{algorithm}
          \end{minipage}}
\end{center}

\vspace{-0.1cm}
\section{Experiments}
\label{sec:experiments}
In this section, first, the experimental dataset is described. Afterward, we explain some details of the experimental-setting, and next, we will present the results based on the metrics discussed above.

\subsection{Dataset}
In our experiments, we used a Brazilian bank's dataset, according to which 3.74\% of all transactions are fraudulent. Nine splits are generated from all transactions in the dataset. Each split has two parts. The first part, which contains 70\% of transactions, is for the training phase. The second part, which contains 30\% of transactions, is for the testing phase. The number of fraudulent and legitimate transactions in each split is shown in \autoref{tab:Freq} \cite{Gadi2008Comparison}. We used MATLAB 2016 software for AIS and ARO implementation. Thus, we changed the format of datasets to CSV.
We trained the fraud detection system by two methods, i.e., ARO and AIS, as explained in the previous sections. 
\subsection{Experimental Settings}
\label{experimentalsettings}
After training the model, in the second step, we ran our model on the validation data with labeled transactions as fraud or normal. Then, in the next step, a comparison is made between the predicted labels and real labels by calculating four parameters:
\begin{itemize}
    \item $False~positive$ $(FP)$: The number of normal transactions that mistakenly predicted as frauds by our method. 
    \item $False~negative$ $(FN)$: The number of fraud transactions that mistakenly predicted as normal by our method. 
    \item $True~positive$ $(TP)$: The number of fraud transactions that correctly detected by our method.
    \item $True~negative$ $(TN)$: The number of normal transactions that correctly detected by our method. 
\end{itemize}
Using the above parameters, we are able to compute some common metrics to evaluate the performance. We used four metrics in our testing phase:
\begin{itemize}
    \item $Sensitivity$ $(\frac{TP}{TP+FN})$:
          It is the ability to recognize a fraudulent transaction as a fraudulent one.
    \item $Precision$ $\frac{TP}{TP+FP}$:
          It is the accuracy on cases predicted as fraud.
    \item $Specificity$ $\frac{TN}{FP+TN}$: 
          It is the ability to recognize a legitimate transaction as a legitimate one.
    \item $Accuracy$ $\frac{TP+TN}{TP+FP+TN+FN}$:
          It presents the proportion of correct predictions. 
\end{itemize}

In addition, we measured training and testing time, which are critical issues in fraud detection. We used Equation \ref{equ:e8} for cost calculating on this dataset. Gadi et al. used this formula because the dataset has 100\% of fraudulent records and only 10\% of legitimate records \cite{Gadi2008Credit}.
\begin{equation}
    \label{equ:e8}
    Cost=100\times{FN}+10\times{FP}+TP
\end{equation}

\begin{figure}[tb]
	\centering
	\includegraphics[width=\textwidth]{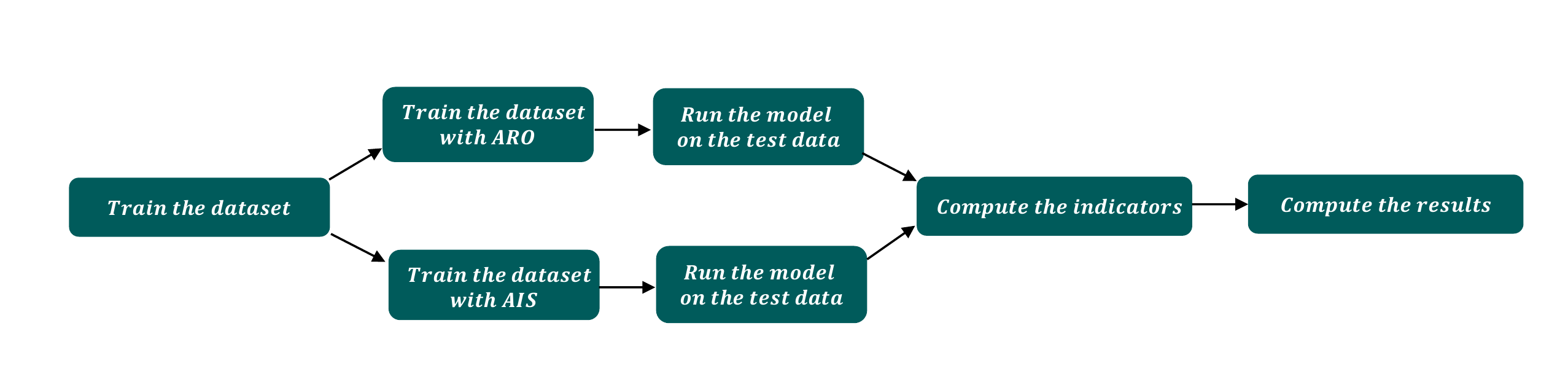}  
	\caption{Fraud detection system. }
	\label{fig:Fraud}
\end{figure}

In the next step, we compared the performance of the two algorithms. The whole process of the fraud detection system is described in \autoref{fig:Fraud}.

As mentioned before, each dataset has a specific cut point. By trial and error method, we found the cut points presented in \autoref{tab:datasets}.
In each training dataset, there are about 28,000 records with 17 features. Finally, test (or validation) datasets are used in the testing phase. For testing the samples obtained by ARO or AIS method, these steps are followed:

\begin{table}[t]
\centering
	\caption{Cut points in the datasets.}
	\begin{tabular}{|l|l|l|l|l|l|l|l|l|l|}
		\hline
		dataset & 1 & 2 & 3 & 4 & 5 & 6 & 7 & 8 & 9 \\ \hline
		Cut point & 0.1754 & 0.1841 & 0.1739 & 0.1762 & 0.175 & 0.1777 & 0.176 & 0.1916 & 0.1749 \\ \hline
	\end{tabular}
		\label{tab:datasets}
\end{table}

\begin{enumerate}
 \item The distance of the record from all normal samples is measured.
 \item The distance is divided by the number of normal samples. One can call it final distance.
  \item	If the final distance is below the best cut-off value, one can categorize the distance as normal, otherwise it would be fraudulent.
\end{enumerate}
The performance is measured by the metrics discussed above. For the AIS method, we have provided the results presented in~\cite{Halvaiee2014novel}, also the results of our implemented version of this algorithm to have a more fair comparison. All the codes are available in https://gitlab.com/Anahita-Farhang/ARO-AIS.

\begin{table}[t]
	
	\caption{Number of fraudulent and legitimate transactions in datasets.}
	\resizebox{\textwidth}{!}{
	\begin{tabular}{|l|l|l|l|l|l|l|l|l|l|l|}
		\hline
		Split type & Transaction type & 1 & 2 & 3 & 4 & 5 & 6 & 7 & 8 & 9 \\ \hline
		\multirow{2}{*}{train} & Legitimate & 27,904 & 28,012 & 28,061 & 28,145 & 28,045 & 27,973 & 28,113 & 27,884 & 28,188 \\ \cline{2-11} 
		& Fraudulent & 1,084 & 1,092 & 1,088 & 1,075 & 1,081 & 1,116 & 1,099 & 1,106 & 1,100 \\ \hline
		\multirow{2}{*}{test} & Legitimate & 12,184 & 12,076 & 12,027 & 11,943 & 12,043 & 12,115 & 11,975 & 12,204 & 11,960 \\ \cline{2-11} 
		& Fraudulent & 475 & 467 & 471 & 484 & 478 & 443 & 460 & 453 & 459 \\ \hline
	\end{tabular}}
\label{tab:Freq}
\end{table}

\subsection{Experimental results}
\begin{figure}[h!]
	\centering
	\includegraphics[width=0.99\textwidth]{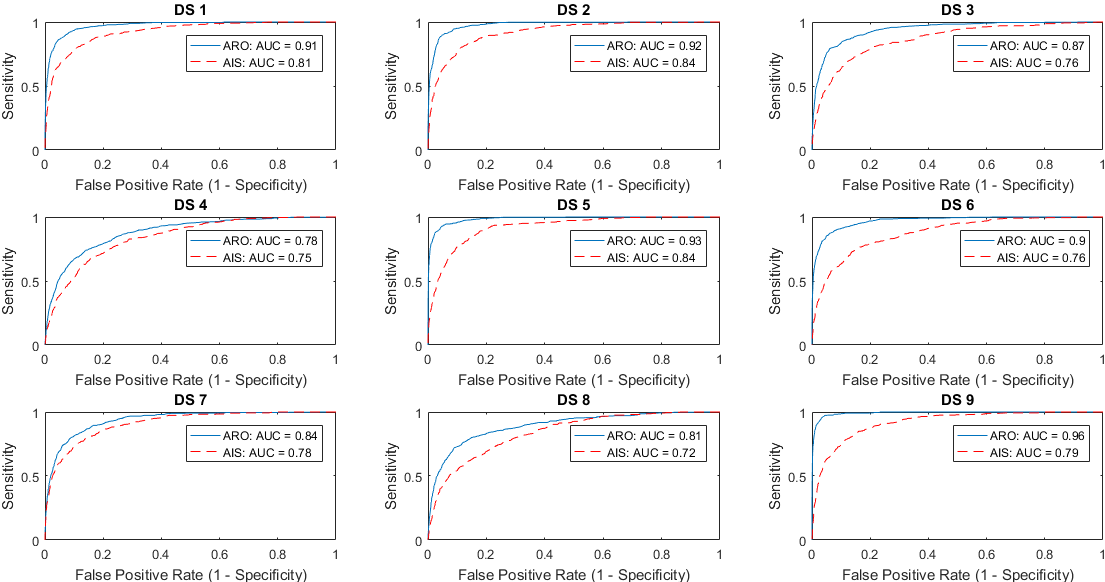}    
	\caption{ROC curve of testing results with ARO and AIS algorithm for all datasets.}
	\label{fig:ROC}
\end{figure}

This subsection presents the computational results of running AIS and ARO algorithms\footnote[1]{All the experiments were performed in a PC with an Intel® Core™ i5-3210M CPU @
2.5GHz with 4GB RAM in Windows 8(x64).}. In sensitivity, precision, specificity, and accuracy, ARO achieved a higher average than AIS, as shown in \autoref{tab:AROAISsen}. For training time, test time, and cost, ARO shows better performance. Results are shown in \autoref{tab:AROAIStime}. As shown in \autoref{fig:ROC}, ROC curve of testing results with ARO and AIS algorithm for all datasets, by implementing ARO, AUC, which is a suitable criterion for imbalanced datasets, is increased by 13\% more than the AIS method. It shows that for each cut-off value, ARO outperforms AIS.

\begin{table}[b]

	\centering
	\caption{ The results of implementing ARO and AIS on datasets.}
	\begin{tabular}{|l|l|l|l|l|l|l|l|l|}
		\hline
		Metric & \multicolumn{2}{l|}{Sensitivity} & \multicolumn{2}{l|}{Precision} & \multicolumn{2}{l|}{Specificity} & \multicolumn{2}{l|}{Accuracy} \\ \hline
		Method & ARO & AIS & ARO & AIS & ARO & AIS & ARO & AIS \\ \hline
		DS 1 & 0.86 & 0.68 & 0.42 & 0.33 & 0.95 & 0.95 & 0.95 & 0.94 \\ \hline
		DS 2 & 0.88 & 0.8 & 0.46 & 0.22 & 0.96 & 0.89 & 0.96 & 0.89 \\ \hline
		DS 3 & 0.79 & 0.61 & 0.34 & 0.22 & 0.94 & 0.92 & 0.93 & 0.9 \\ \hline
		DS 4 & 0.65 & 0.63 & 0.23 & 0.16 & 0.91 & 0.87 & 0.9 & 0.86 \\ \hline
		DS 5 & 0.88 & 0.78 & 0.58 & 0.23 & 0.97 & 0.9 & 0.97 & 0.89 \\ \hline
		DS 6 & 0.86 & 0.58 & 0.38 & 0.24 & 0.95 & 0.94 & 0.95 & 0.92 \\ \hline
		DS 7 & 0.74 & 0.6 & 0.32 & 0.34 & 0.94 & 0.96 & 0.93 & 0.94 \\ \hline
		DS 8 & 0.72 & 0.51 & 0.23 & 0.2 & 0.91 & 0.93 & 0.91 & 0.91 \\ \hline
		DS 9 & 0.95 & 0.63 & 0.54 & 0.33 & 0.97 & 0.95 & 0.97 & 0.94 \\ \hline
		\textbf{Average} & \textbf{0.81 }& \textbf{0.65} & \textbf{0.39 }& \textbf{0.25} & \textbf{0.95} & \textbf{0.92} & \textbf{0.94} & \textbf{0.91} \\ \hline
	\end{tabular}
	\label{tab:AROAISsen}
\end{table}

\begin{table}[tb]
	\centering
	\caption{ The results of implementing ARO and AIS on datasets.}
	\begin{tabular}{|l|l|l|l|l|l|l|l|l|}
		\hline
		\multicolumn{1}{|c|}{\textbf{Metric}} & \multicolumn{2}{c|}{\textbf{Train time (s)}} & \multicolumn{2}{c|}{\textbf{Test time (s)}} & \multicolumn{2}{c|}{\textbf{Cost}} & \multicolumn{2}{c|}{\textbf{AUC}} \\ \hline
		Method & ARO & AIS & ARO & AIS & ARO & AIS & ARO & AIS \\ \hline
		DS 1 & 8.08 & 24.25 & 1.87 & 1.85 & 12,570 & 22,072 & 0.91 & 0.81 \\ \hline
		DS 2 & 8.46 & 24.90 & 1.32 & 1.19 & 10,781 & 23,213 & 0.92 & 0.84 \\ \hline
		DS 3 & 7.33 & 24.65 & 2.22 & 1.75 & 17,473 & 28,966 & 0.87 & 0.76 \\ \hline
		DS 4 & 4.68 & 24.44 & 1.23 & 1.52 & 27,923 & 33,864 & 0.78 & 0.75 \\ \hline
		DS 5 & 4.63 & 24.66 & 1.13 & 1.62 & 9,132 & 23,115 & 0.93 & 0.84 \\ \hline
		DS 6 & 7.78 & 24.57 & 1.34 & 1.75 & 12,889 & 26,925 & 0.9 & 0.76 \\ \hline
		DS 7 & 3.8 & 24.25 & 1.27 & 1.68 & 19,522 & 24,224 & 0.84 & 0.78 \\ \hline
		DS 8 & 7.06 & 24.32 & 1.79 & 1.72 & 23,944 & 31,599 & 0.81 & 0.72 \\ \hline
		DS 9 & 4.43 & 25.30 & 1.54 & 1.80 & 6,407 & 23,071 & 0.96 & 0.79 \\ \hline
		\textbf{Average} & \textbf{6.25} & \textbf{24.59} & \textbf{1.52} & \textbf{1.65} & \textbf{15,627} & \textbf{26,339} & \textbf{0.88} & \textbf{0.78} \\ \hline
	\end{tabular}
	\label{tab:AROAIStime}
\end{table}

Finally, two non-parametric statistical tests (i.e, wilcoxon and Kruskal-Wallis) were conducted to ensure the statistical significance in terms of accuracy for the ARO model. The Wilcoxon test results are illustrated in \autoref{tab:Wilcoxon} that the ARO model almost reaches the significance level compared to AIS. The Kruskal-Wallis test was used to show the equality of results in all nine sections of the dataset. Results are presented in \autoref{tab:Kruskal}.

\section{Discussion}
\label{sec:disc}
We trained the system with ARO and AIS methods. As mentioned in \autoref{sec:back}, given that ARO is a single-solution evolutionary algorithm, it responds faster than the AIS that generates a community of data \cite{farasat2010aro,mansouri2011aro,salmeron2019learning}. Therefore, the good speed with no parameter setting and good convergence rate have made ARO a good candidate versus AIS, and in our experiment, this claim was confirmed in four indicators.
In classification problems, there are some common metrics to evaluate the performance: sensitivity (recall), precision, specificity, and accuracy. These four metrics have been measured in our testing phase.
AUC was measured, which is the area under the ROC curve. ROC curve plots sensitivity versus false-positive rate. In fact, the cut-off value in the test phase is located at the top left corner in ROC curve. It is the point where sensitivity and specificity are equal.
Gadi et al. found that if they use a cost function shown in Equation \ref{equ:e8} in which they adopted an average cost of 1 dollar for every verification and an average loss of 100 dollars for every undetected fraud, they will obtain more applicable results. They used this formula because the dataset has 100\% of fraudulent records and only 10\% of legitimate records \cite{Gadi2008Credit, Halvaiee2014novel,akila2018cost,ghobadi2016cost}. This was considered to be more similar to the practice used for a fraud score compared to a ROC curve that compares multiple references simultaneously \cite{Gadi2008Credit}.

One of the main problems of AIS is the extreme need for a hyper-parameter setting, which is not present in ARO. ARO is a bio-inspired algorithm, and we aimed to test this algorithm against one of the algorithms that works best in detecting fraud on a Brazilian bank's dataset. Compared to other studies on this dataset, the detecting speed and the computational cost were important. 
We trained each dataset by each algorithm thirty times and registered the results of the best cost. As shown in Tables \ref{tab:AROAISsen} and \ref{tab:AROAIStime}, ARO has better performance than AIS in all the metrics. The ARO method's best performance appears on the ninth dataset with the sensitivity of 0.95 and the cost of 6,407, which is better than the AIS method (sensitivity=0.63 \& cost=23071). The ARO method's worst performance appears on the fourth dataset with the sensitivity of 0.65 and the cost of 27,923, which is still better than the AIS method (sensitivity{=}0.63  \& cost{=}33864). The average sensitivity for ARO is 0.81 with the average precision 0.39. For the AIS technique, the average sensitivity and precision are 0.65 and 0.25. Training time, which is a vital issue in fraud detection, has been remarkably reduced. The average training time for ARO is 6.25s, so ARO fraud detection system can be considered as a real-time one. 
ARO improved sensitivity up to 25\%, and precision up to 56\%, decreased cost up to 41\%, and training time up to 75\%.
The first fraud detection on our dataset was implemented by Gadi et al. He proved that by optimizing the parameters, AIS is the best method in comparison with BN, NB, and DT \cite{Gadi2008Credit}. One of the best fraud detection systems on this dataset was performed in \cite{Halvaiee2014novel}. They employed AFDM which is a kind of improved AIS method. They achieved 17,389 for cost and 79 seconds for training time. By implementing ARO, we achieved 15,627 for cost and 6.25 for training time which are considerably better than the previous results. The obtained results and the results of the previous researches are shown in \autoref{tab:history}. 
\begin{table}[]
\centering
	\caption{History of the previous and obtained results.}
	\begin{tabular}{|l|l|l|l|l|}
		\hline
		Method & AIS & AFDM & AIS & ARO \\ \hline
		Cost & 23,303 & 17,389 & 26,339 & 15,627 \\ \hline
		Reference & \cite{Gadi2008Credit} & \cite{Halvaiee2014novel} & Proposed AIS & Proposed ARO \\ \hline
	\end{tabular}
	\label{tab:history}
\end{table}

In the last part, we have used two non-parametric statistical tests. We have applied Wilcoxon to show the significant difference between AIS and ARO algorithms. This test ranks all differences and applies a negative sign to all the ranks where the difference between the two observations is negative. The hypothesis $H0$ in this test is the equality of the two algorithms and, as shown in Table \ref{tab:Wilcoxon}, in accuracy, sensitivity, precision, train time, and cost, because of the p-values which are less than alpha ($\alpha = 0.05$), this hypothesis was rejected that means two algorithms are not equal. Also, in the Wilcoxon test, negative rank for train time, test time, cost, and positive rank for other indices showed that ARO performs better than AIS in all indices.

\begin{table}[!]

	\caption{ Results of Wilcoxon signed-rank test with $\alpha{=}0.05$.}
	\resizebox{\textwidth}{!}{
	\begin{tabular}{|l|l|l|l|l|l|l|l|l|}
		\hline
		Compared model & Sensitivity & Precision & Specificity & Accuracy & Train time (s) & Test time (s) & Cost & AUC \\ \hline
		Asymp. Sig. & 0.008 & 0.011 & 0.118 & 0.020 & 0.008 & 0.314 & 0.008 & 0.008 \\ \hline
	\end{tabular}}
	\label{tab:Wilcoxon}
\end{table}

We have done Kruskal-wallis test because our dataset was divided into nine sections and it is important to check whether all the samples are originated from the same distribution. We performed this test to check whether there is a significant difference between the nine samples in each index. The results are shown in Table \ref{tab:Kruskal}. The hypothesis $H0$ in this test is the equality between all the nine samples. Due to the p-values which are greater than alpha ($\alpha = 0.05$), and also because of the values of chi-square that are 8, which is less than 15.5073 ($\chi^2_{0.05}=15.5073$ with $df=8$), the $H0$ hypothesis cannot be rejected which means in all indexes, our nine sections are the same.

\begin{table}[!]

	\caption{Results of Kruskal-Wallis test.}
	\resizebox{\textwidth}{!}{
	\begin{tabular}{|l|l|l|l|l|l|l|l|l|}
		\hline
		Compared model & Sensitivity & Precision & Specificity & Accuracy & Train time (s) & Test time (s) & Cost & AUC \\ \hline
		Chi-square & 8.000 & 8.000 & 8.000 & 8.000 & 8.000 & 8.000 & 8.000 & 8.000 \\
		\hline
		df & 8 & 8 & 8 & 8 & 8 & 8 & 8 & 8 \\\hline
		Asymp. Sig. & 0.433 & 0.433 & 0.433 & 0.433 & 0.433 & 0.433 & 0.433 & 0.433 \\ \hline
	\end{tabular}}
		\label{tab:Kruskal}
\end{table}

As it was discussed, we have achieved promising results by using ARO algorithm. However, there is room for improvement. For example, an algorithm can be employed to choose the optimized cut-point values in Section \ref{experimentalsettings}. Moreover, to increase the performance of the algorithm, we suggest using cloud computing, i.e. implementing ARO algorithm on a cloud-based file system (e.g, Hadoop) which makes data parallelization possible. Furthermore, new methods in the deep learning area show progress in terms of time in comparison with metaheuristic algorithms. Therefore, employing deep learning methods may reduce the training time and have positive impacts on the final results.

\section{Conclusion}
\label{sec:conc}
Fraud is a critical concern for financial services (e.g., commercial banks, investment banks, insurance companies, etc.) and individuals. Different types of fraud cost millions of dollars every year. Among different types of fraud, credit card fraud
is the most common one and several solutions have been proposed to detect fraudulent transactions. In this paper, we have implemented the ARO (Asexual Reproduction Optimization) in credit card fraud detection. This effective approach has achieved better results than the best techniques implemented on our dataset so far. We have compared the results with those of the AIS, which was one of the best methods ever implemented on the benchmark dataset. 

The chief focus of the fraud detection studies is finding the algorithms that can detect legal transactions from the fraudulent ones with high detection accuracy in the shortest time and at a low cost. ARO meets all these demands.
ARO is an Evolutionary Single-Objective Optimization algorithm with lots of advantages that make it suitable for implementing in fraud detection problems. First of all, being an individually based technique, it converges faster to the global optimal point. Secondly, it has good exploration and exploitation rates. Thirdly, it has no parameter settings, which is a common issue in meta-cognitive problems such as Genetic Algorithms (GA), Annealing Simulation (SA), Taboo Search (TS), and Particle Swarm Optimization (PSO).
Results show that ARO has increased the AUC, sensitivity, precision, specificity, and accuracy by 13\%, 25\%, 56\%, 3\%, and 3\%, in comparison with AIS, respectively. We have achieved a high precision value indicating that if ARO detects a record as a fraud, with a high probability, it is a fraud one. Supporting a real-time fraud detection system is another vital issue. ARO outperforms AIS not only in the mentioned criteria, but also decreases the training time by 75\% in comparison with the AIS, which is significant. Furthermore, two non-parametric statistical tests (i.e., Wilcoxon and Kruskal-Wallis) were conducted to ensure the statistical significance in terms of accuracy for the ARO model. The Wilcoxon test results show that the ARO model almost reaches the significance level compared to AIS. The Kruskal-Wallis test was used to show the equality of results in all nine sections of the dataset. The results of applying these two statistical tests ensure the statistical significance in our study.

\section{Future Work}
\label{sec:furturework}
Our framework has addressed the problems such as high costs and training time in credit card fraud detection. Although, there is still room for further improvement. To increase the performance of the proposed method, it is possible to test the proposed model in a cloud environment, i.e., Hadoop. Moreover, ARO can be compared to PSO and QPSO, which have fewer parameter settings than AIS. 

In addition, the writers believe
ARO has the potential to obtain much better results. One improvement can be done by weighting the fields that compose a transaction. In fact, there are plenty of fields in a transaction and some fields are more important than other fields. Therefore, we can increase or decrease the effect of the field on the final results through weighting fields in the distance function. Furthermore, the distance function can be different for each property in the dataset. As we discussed, each transaction has several fields with different meanings. Then the concept of distance is not the same for all the fields. As an example, suppose the person goes shopping once per month. So the distance of 30 is usual and it equals zero for the time concept. However, the distance of 30 for the amount column is important and it is not equal to zero. Therefore, considering application-based distance functions for each field is an interesting point to address in future work.

\bibliography{bibliography}

\end{document}